\documentclass[10pt,twocolumn,letterpaper]{article}

\usepackage{wacv}              

\usepackage[accsupp]{axessibility} 
\usepackage{graphicx}
\usepackage{amsmath}
\usepackage{amssymb}
\usepackage{booktabs}
\usepackage{bm}

\usepackage[pagebackref,breaklinks,colorlinks]{hyperref}

\usepackage[capitalize]{cleveref}
\crefname{section}{Sec.}{Secs.}
\Crefname{section}{Section}{Sections}
\Crefname{table}{Table}{Tables}
\crefname{table}{Tab.}{Tabs.}

\def\wacvPaperID{529} 
\def\confName{WACV}
\def\confYear{2025}

\begin{document}

\title{WARLearn: Weather-Adaptive Representation Learning}

\author{Shubham Agarwal \quad Raz Birman \quad Ofer Hadar \\
Ben-Gurion University of the Negev, Israel \\
{\tt\small agarwals@post.bgu.ac.il, raz.birman@gmail.com, hadar@bgu.ac.il}
}
\maketitle

\begin{abstract}
This paper introduces WARLearn, a novel framework designed for adaptive representation learning in challenging and adversarial weather conditions. Leveraging the in-variance principal used in Barlow Twins, we demonstrate the capability to port the existing models initially trained on clear weather data to effectively handle adverse weather conditions. With minimal additional training, our method exhibits remarkable performance gains in scenarios characterized by fog and low-light conditions. This adaptive framework extends its applicability beyond adverse weather settings, offering a versatile solution for domains exhibiting variations in data distributions. Furthermore, WARLearn is invaluable in scenarios where data distributions undergo significant shifts over time, enabling models to remain updated and accurate. Our experimental findings reveal a remarkable performance, with a mean average precision (mAP) of 52.6\% on unseen real-world foggy dataset (RTTS). Similarly, in low light conditions, our framework achieves a mAP of 55.7\% on unseen real-world low light dataset (ExDark). Notably, WARLearn surpasses the performance of state-of-the-art frameworks including FeatEnHancer, Image Adaptive YOLO, DENet, C2PNet, PairLIE and ZeroDCE, by a substantial margin in adverse weather, improving the baseline performance in both foggy and low light conditions. The WARLearn code is available at \url{https://github.com/ShubhamAgarwal12/WARLearn}
\end{abstract}

\section{Introduction}
\label{sec:intro}

One of the main limitations of computer vision based algorithms in the real-world setting is their inability to adapt to changing weather conditions. 
Visibility changes due to extreme weather and lighting conditions pose a significant challenge in tasks like object detection. This challenge is particularly relevant in scenarios such as autonomous cars, where inaccuracies in object recognition might lead to loss of life and property. Lately, there have been several advancements in this field. One approach that has received a lot of attention involves cleaning of the image to reduce the distortion caused by extreme weather. This includes a pre-processing step that is done using traditional image processing techniques \cite{PreProcess4, PreProcess2, PreProcess3, PreProcess1} or using deep learning based approaches \cite{c2pnet, Zero-DCE, pairlie, featenhancer}. Hashmi \etal \cite{featenhancer} try to learn the enhancement module weights based on the downstream task loss. Some recent methods also include additional pre-processing layers in the detection model itself \cite{YoloAdavtive, denet} that learn to remove the weather specific degradation. These additional steps increase the computational complexity and also the latency. Some other methods try to jointly learn image enhancement and object recognition \cite{AdverseWeather, li2023detection, ding2023cf} which makes it difficult to tune the model parameters and results in loss of performance. MAET \cite{MAET} leverages adversarial learning to correctly detect objects in images modified by exposure transformations, while a generator tries to create images that challenge the detection network. There are also approaches that employ domain adaptation principles \cite{domainadapt1, domainadapt2, domainadapt3, domainadapt4}. Sindagi \etal \cite{domainadapt1} do this by separating the image style and weather specific features. The model then uses self supervised contrastive learning to learn features robust to adverse weather. DA-YOLO \cite{domainadapt2} also separates the features into image style and object specific. These are then separately regularised for adversarial training. Sindagi \etal \cite{domainadapt3} proposes cleaning the feature space by using prior-adversarial loss based on weather specific information. Li \etal \cite{domainadapt4} also adopts different alignment techniques for image and object level features.

\begin{figure}[t]
  \centering
   \includegraphics[width=1\linewidth]{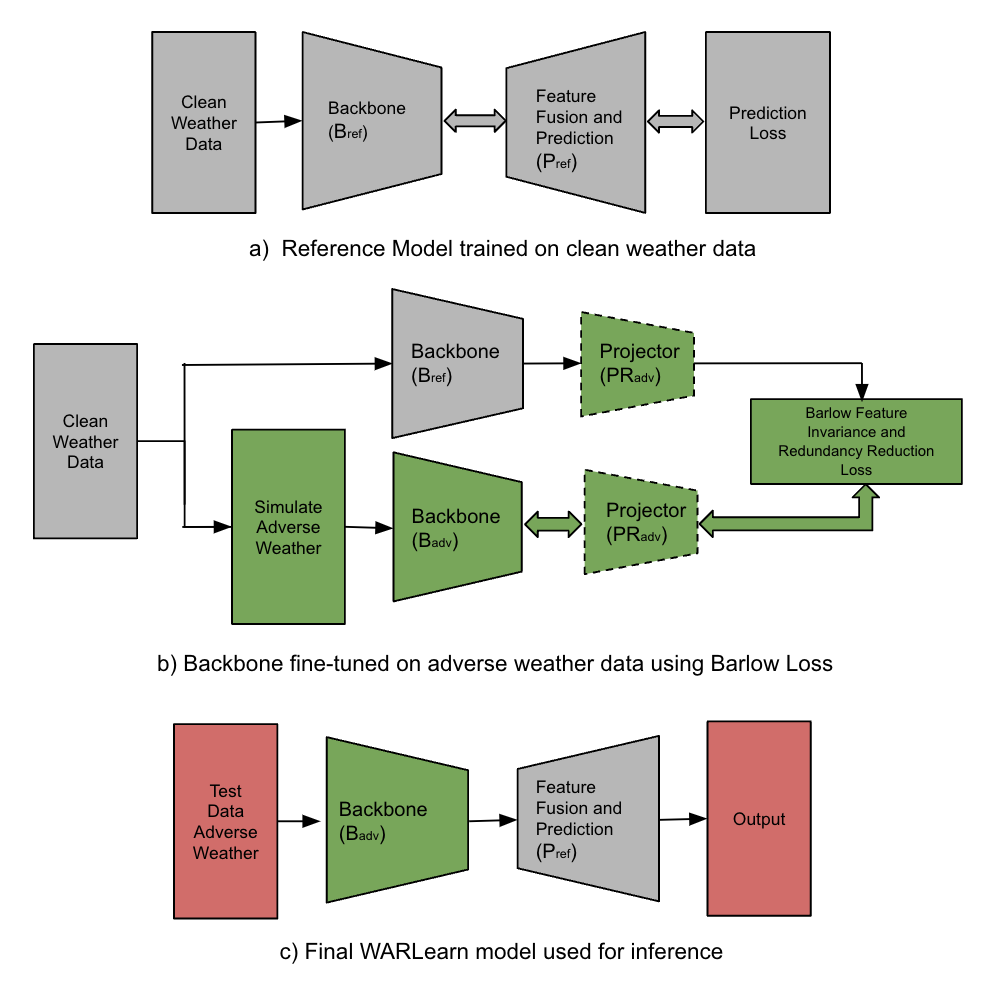}

   \caption{Our proposed WARLearn framework initiates with the training of a model using clean weather data (a). Subsequently, leveraging the Barlow feature in-variance and redundancy reduction loss, we try to align the feature representations of adverse weather conditions with those obtained under clear weather conditions (b). This step enables the creation of a hybrid model, where the backbone is derived from the refined feature representations, while feature fusion and prediction components stem from the initial clean weather data (c). This hybrid model is then employed for robust predictions in adversarial weather conditions. In the figure, double-sided thick arrows indicate the flow of information between modules undergoing training in the respective stage. Dotted lines represent shared weights.}
   \label{fig:warlearn}
\end{figure}

WARLearn proposes a novel representation learning framework to adapt existing models trained on clean data to adverse weather conditions. Building on the idea of distribution shift in extreme weather, we learn feature representations that account for this shift. \Cref{fig:warlearn} provides a flowchart explaining the WARLearn framework. In WARLearn, we first train the model on clean weather data. Subsequently, we extract and store the representations from the feature extraction part of the network corresponding to each clean image. Following this, we induce extreme weather conditions onto the clean data, generating a set of synthetic images under extreme weather scenarios corresponding to their clean counterparts.  We then fine tune just the feature extractor part of the network on these synthetic extreme weather images, matching their feature representations with that of the corresponding clean images. This fine-tuning operation is facilitated through the Barlow Twins Loss \cite{BarlowTwins}. The final model for inference is obtained by combining the feature extractor obtained after the fine tuning step with the prediction module obtained from the clean weather training.

In contemporary object detection research focused on challenging weather conditions, most of the latest studies \cite{YoloAdavtive, denet, Pami1} build upon the YOLOv3 \cite{yolov3} framework. For a fair assessment, we also adopt the YOLOv3 object detector within the WARLearn framework.  We rigorously assess the performance of WARLearn on both simulated and previously unseen real-world data. Notably, with minimal fine-tuning and  without any added complexity, our model either outperforms existing methods or closely approaches the state-of-the-art benchmarks, when confronted with unseen extreme weather real-world data. We provide a comprehensive comparative analysis, specifically focusing on scenarios characterized by foggy and low-light weather conditions. We also provide experimental results using the WARLearn framework with YOLOv8 \cite{Yolov8}, demonstrating the versatility and effectiveness across different neural network architectures.

\section{Related Work}

\textbf{Object Detection in Adverse Weather:}
Recently, there has been a lot of focus on vision algorithms in adverse weather conditions due to direct applicability in the real world. One of the primary research direction in this field is using the image enhancement methods or neural network based preprocessing to try and remove the effect of adverse weather \cite{PreProcess4, PreProcess2, PreProcess3, PreProcess1, Adw, Adw2, preprocess11, preprocess14, preprocess15}. Some of them solely focus on defogging \cite{defogging1, preprocess13}. C2PNet \cite{c2pnet} proposes a dehazing network combining the atmospheric scattering model and curricular contrastive regularization. Zero-DCE \cite{Zero-DCE} uses non-reference loss functions to predict the pixel dynamic range adjustment for low-light image enhancement. Hahner \etal \cite{Foggy1} proposed a method for simulating foggy weather on LiDAR-based 3D data. Training with this data significantly improves the 3D object detection in foggy weather. Another method that has been explored is having additional layers for preprocessing and training these together with the object recognition \cite{YoloAdavtive}. 

Valanarasu \etal \cite{Transweather} uses a transformer based architecture for image restoration.  Qin \etal \cite{denet} proposed a preprocessing module called DENet that relies on the hypothesis that weather related information is more prominent in the low frequency component of image. This information is then used to refine the high frequency component of image that have features like edges and texture. There have also been attempts to use adversarial learning based feature calibration modules \cite{AdverseWeather}. Huang \etal \cite{Pami1} introduces DSNet trained to learn tasks of visibility enhancement, object classification and object localization simultaneously.
Michaelis \etal \cite{style1} used style transfer to mitigate the effects of adverse weather.

There are several works on using representation learning in adverse weather \cite{represent1, represent2, yuweather, pointclouddenoising2020}. Wu \etal \cite{represent1} proposed a method that uses contrastive learning to learn feature representations that are then passed to the object recognition module. The representations improve the robustness of the model in adverse weather. Yurui \etal \cite{yuweather} try to learn weather specific and weather general features from the representations. These two kind of features are then handled differently in the model for image restoration. Sindagi \etal \cite{domainadapt1} proposes cleaning the feature space by using prior-adversarial loss based on weather specific information. 

\noindent
\textbf{Barlow Twins Loss:}
 Barlow Twins Loss \cite{BarlowTwins} has been used to learn feature representations that are robust to minor distortions in the input. In this method, we first compute the correlation matrix \textit{C} from the features obtained by two distorted versions of the same input. Let $Z^{A}$ and $Z^{B}$ be the two feature vectors, mean centered along batch dimension.

\begin{equation}
  C_{ij} \triangleq \dfrac{\sum_{b}z_{b,i}^{A}z_{b,j}^{B}} {\sqrt{\sum_{b}(z_{b,i}^{A})^2}\sqrt{\sum_{b}(z_{b,j}^{B})^2}}
  \label{eq:eq1}
\end{equation}

where b is the index for batch samples and \textit{i,j} are the indexes along the feature dimension. Now, the Barlow Twins loss is defined as follows:
\begin{equation}
  L_{BT} \triangleq \sum_{i}(1-C_{ii})^2 + \lambda\sum_{i}\sum_{j{\neq}i}C_{ij}^2
  \label{eq:eq2}
\end{equation}

where the first sum in the equation is the \textit{invariance term} that tries to make the diagonal elements of the cross correlation matrix to 1. In doing so, it forces the network to learn distortion invariant features. The second sum, known as \textit{redundancy reduction term} tries to make all the off diagonal elements to 0. This decorrelates different features and helps in redundancy reduction among the features. \textit{$\lambda$} is a positive constant used to balance the importance of the \textit{invariance term} and the \textit{redundancy reduction term}.

Barlow Twins Loss has been recently used in adaptation frameworks alongside task-specific losses. Yassine \etal \cite{blackbox} use the Barlow Twins based ResNet50 feature encoder to get visual features. These features are then aligned with text features.  Dripta \etal \cite{barlowadaptPE} applied it to pose estimation, involving both feature and prediction space adaptation along with regularization, where both the teacher and student models are modified during training. Similarly, Victor \etal \cite{barlowadaptAR} used it for action recognition with an action classification loss, training both source and target domain data on the same backbone. WARLearn differs from these approaches in two key aspects:
\\
1. It uses a fixed teacher backbone as a reference, adapting only the student backbone for adverse condition.
\\
2. It solely relies on Barlow Twins Loss for adaptation, without incorporating any task-specific losses.

\section{WARLearn}
The foundational idea behind our WARLearn approach centers on the premise that clear weather conditions act as a reference for object detection tasks in adverse weather conditions. When the same scene is encountered in adverse weather, performance metrics are expected to deviate from this benchmark. By training the model to discern the distribution shift from clear to adverse weather scenarios, it gains the capability to predict features for the adverse weather image as if the conditions were clear.

\begin{figure}[h]
  \centering
   \includegraphics[width=1\linewidth]{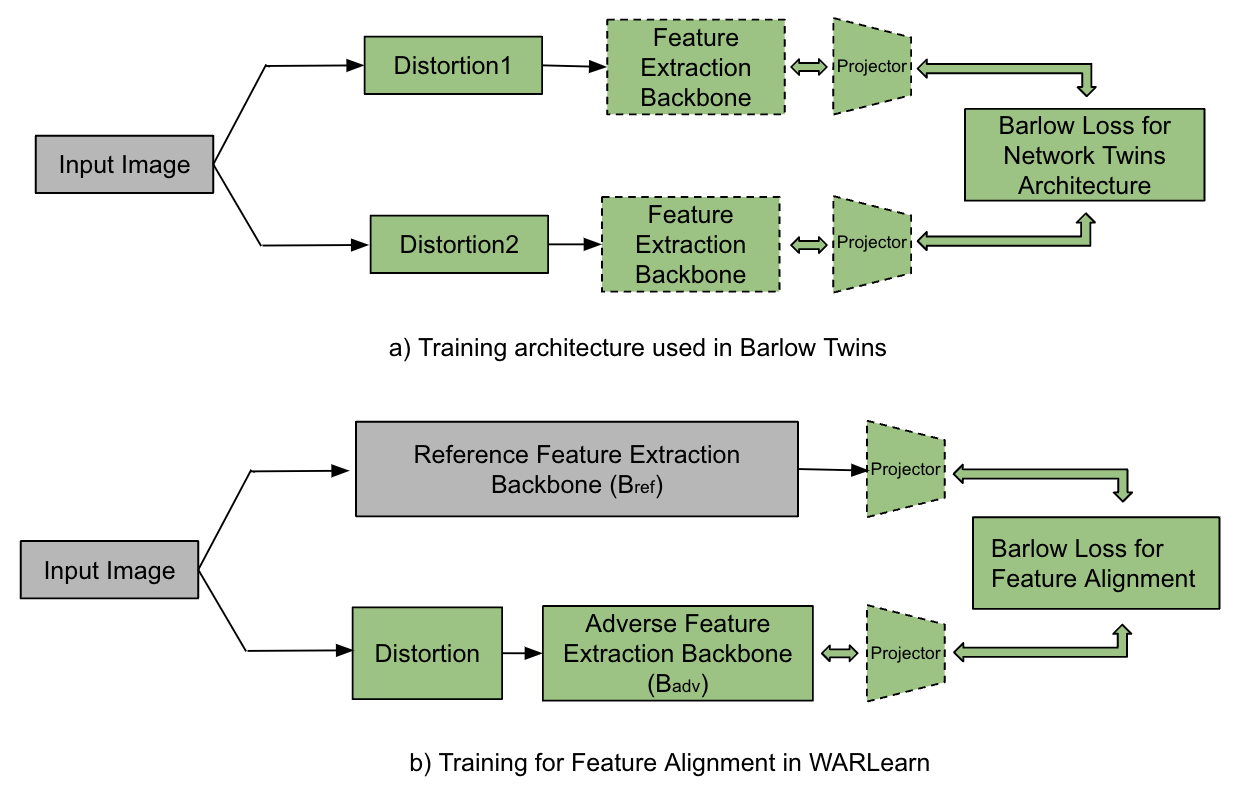}

   \caption{In the Barlow Twins approach, we input two distorted versions of an image into the same feature extractor. This extractor is trained to align the features of the two distorted versions, fostering similarity in their representations. WARLearn employs a foundational feature extractor initially trained on clean data as a reference. It then trains a secondary feature extractor dedicated to distorted images, utilizing the foundational extractor as a reference. The objective is to match the features extracted from distorted images with those from corresponding clean images, thereby learning to mitigate the impact of adverse weather-induced distortions. The modules that undergo training using the Barlow redundancy reduction loss are highlighted in green colour. Broken boundaries indicate weight sharing between identical modules.}
   \label{fig:bt}
\end{figure}

We start by training the model on a clear weather dataset, as illustrated in Figure \ref{fig:warlearn} (a). Let's call this model $M_{ref}$, with its backbone denoted as $B_{ref}$ and the predictor part as $P_{ref}$. Once the clear weather training is complete, we create a copy of $B_{ref}$, named $B_{adv}$, which is further fine-tuned for adverse weather scenario. The fine-tuning process works by first passing the clear weather image through $B_{ref}$ and the corresponding adverse weather image through $B_{adv}$ to obtain the reference features and adverse weather features, respectively. Both these feature sets are projected through a projection network $PR_{adv}$ to reduce the dimensionality. Finally, we compute the Barlow loss between the projected reference features ($Z^{ref}$) and the projected adverse weather features ($Z^{adv}$). The loss is used to train the backbone $B_{adv}$ for adverse weather. This setup is depicted in Figure \ref{fig:warlearn} (b). With this feature alignment setup, Equation \ref{eq:eq1} can be re-written as follows.

\begin{equation}
  C_{ij} \triangleq \dfrac{\sum_{b}\bm{z_{b,i}^{ref}}z_{b,j}^{adv}}{\sqrt{\sum_{b}(\bm{z_{b,i}^{ref}})^2}\sqrt{\sum_{b}(z_{b,j}^{adv})^2}}
  \label{eq:eq3}
\end{equation}

Notice that we now have a constant feature vector corresponding to each image in the clear weather dataset. Since the clear weather backbone $B_{ref}$ remains fixed, the adverse weather features with varying levels of distortions are always compared against the same clear weather feature counterpart $Z^{ref}$. Assuming a single sample in a batch, the diagonal element in the correlation matrix $C_{ii}$ becomes the dot product of the feature set from the clear image and the adverse weather image. The invariance term in the Barlow Twins loss becomes as follows:

\begin{equation}
  L_{invariance} = (1 - \bm{Z^{ref}} \cdot Z^{adv})^2
  \label{eq:eq4}
\end{equation}

As we can see in Equation \ref{eq:eq4}, the invariance loss term is minimized to make the dot product $Z^{ref} \cdot Z^{adv}$ close to 1, thus introducing similarity between the adverse weather image features through $B_{adv}$ and the clear weather features through $B_{ref}$. This ensures that the weights of $B_{adv}$ learn to mitigate the effect of adverse conditions and produce features as if the input image were clear and passed through $B_{ref}$. The redundancy reduction term in Equation \ref{eq:eq2} is used to introduce decorrelation among the feature set, which helps in generalization. Figure \ref{fig:bt}
highlights the difference between the Barlow Twins and the WARLearn framework.

The final model for inference is depicted in Figure \ref{fig:warlearn}(c), where the backbone $B_{adv}$ is obtained from step Figure \ref{fig:warlearn} (b) to mitigate the effect of adverse weather, and the $P_{ref}$ module remains the same as that in Figure \ref{fig:warlearn} (a). This setup has inherent modularity, and the same $P_{ref}$ module can be utilized for multiple adversarial scenarios by plugging in the corresponding $B_{adv}$.

The WARLearn framework can work under two conditions. Firstly, the model needs to have separable feature extraction and prediction modules. Secondly, we need a simulation algorithm to apply the desired distortions on the clear image dataset so that there is image and feature correspondence for Barlow loss based feature alignment. With these conditions satisfied, our framework can be extended easily to other adverse  scenarios.

\section{Experiment Setup}
We conducted experiments using the WARLearn framework on the YOLOv3 model with the Darknet-53 backbone \cite{yolov3}. Darknet-53 is a sequential, 53 layer deep network that features efficient feature extraction, residual connections, and multi-scale feature integration. The Darknet-53 module serves as our feature extractor. 

The base YOLOv3 model with clean data was trained on the widely used Pascal VOC dataset \cite{pascal-voc-2007, pascal-voc-2012}, which comprises 20 classes: \textit{aeroplane, bicycle, bird, boat, bottle, bus, car, cat, chair, cow, dining-table, dog, horse, motorbike, person, potted-plant, sheep, sofa, train, tv-monitor}. The dataset contains 19,561 training images and 4,952 test images. For training, we utilized a batch size of 32 and the ADAM optimizer \cite{Adam} with an initial learning rate of 0.02. The experiments were set up using PyTorch, and we employed an NVIDIA GEFORCE RTX 3090, 32 GB GPU for execution. The clean data training was performed for 600 epochs.

To simulate adverse weather conditions, specifically foggy weather and low light, on the PascalVOC dataset, we followed the methodology described in \cite{AdverseWeather}. Foggy images were generated using the atmospheric scattering model developed by \cite{fog1, fog2}, where:

\begin{equation}
  I(x) = J(x)t(x) + A (1-t(x))
  \label{eq:fog1}
\end{equation}

Here, \(I(x)\) is the foggy image, \(J(x)\) is the clean image, \(A\) is the global atmospheric light, and \(t(x)\) is the medium transmission map obtained using the scattering coefficient of the atmosphere \(\beta\) and scene depth \(d(x)\):

\begin{equation}
  t(x) = e^{-\beta}d(x)
  \label{eq:fog2}
\end{equation}

\begin{equation}
  d(x) = -0.04\rho + \sqrt{\max(\text{rows}, \text{cols})}
  \label{eq:fog3}
\end{equation}

Here, \(\rho\) is the distance of the current pixel from the center, and \(\text{rows}\) and \(\text{cols}\) are the number of rows and columns in the image, respectively. We set \(A\) to 0.5 and 
\begin{equation}
\beta = 0.01i + 0.05
\label{eq:beta}
\end{equation}We vary \(i\) from 0 to 9 to obtain 10 levels of foggy images for each clean image in the PascalVOC dataset.

Similarly, for obtaining low-light images, we used the transformation:

\begin{equation}
  l(x) = x^{\gamma}
  \label{eq:lowlight1}
\end{equation}

Here, \(x\) is the intensity of the input pixel, and \(\gamma\) is randomly sampled from a uniform distribution within the range [1.5, 5]. We randomly sampled 10 different \(\gamma\) values to obtain 10 different low-light images corresponding to each image in the PascalVOC dataset.

In the original Barlow Twins paper \cite{BarlowTwins}, experiments show that the loss (Equation \ref{eq:eq2}) is not very sensitive to the hyper-parameter $\lambda$. We adopt the $\lambda$ value of 0.001 for all our experiments. To highlight the significance of the redundancy reduction term, we include a comparison of WARLearn with $\lambda = 0$ in the ablation studies.
 
During fine-tuning for adverse weather, we reduced the learning rate to 0.0002 and fine-tuned the backbone for 10 epochs. To evaluate performance on real-world data, we used the RTTS \cite{RTTS} and ExDark \cite{Exdark} datasets. It's important to note that these datasets were not used during training or fine-tuning. \Cref{tab:classes} lists the common classes between these datasets and the PascalVOC dataset. Note that we train all the WARLearn models with all the 20 classes in PascalVOC dataset.

\begin{table*}[h]
  \centering
  \begin{tabular}{p{4.9cm}p{2cm}p{1cm}p{0.8cm}p{1.2cm}p{5.1cm}}
    \toprule
    \textbf{Dataset} & \textbf{Type} & \textbf{Images} & \textbf{Classes} & \textbf{Instances} & \textbf{Purpose} \\
    \toprule
    PascalVOC train & Real & 19561 & 20 & 40058&Clear weather training/Part of mixed \\
    PascalVOC train foggy& Simulated & 195610 & 20 & 400580&Part of foggy mixed \\
    PascalVOC train lowlight& Simulated & 195610 & 20 & 400580& Part of lowlight mixed \\
    PascalVOC train foggy mixed& Real+Simulated & 215171 & 20 & 440638&Foggy weather training \\
    PascalVOC train lowlight mixed& Real+Simulated & 215171 & 20 & 440638& Lowlight weather training \\
    \midrule
    PascalVOC test & Real & 4952 & 20 & 12032 & Clear weather testing \\
    PascalVOC test foggy (SimFoggy)& Simulated & 49520 & 20 & 120320 & Foggy weather testing \\
    PascalVOC test lowlight (SimLL) & Simulated & 49520 & 20 & 120320 & Lowlight weather testing \\
    RTTS& Real & 4322 & 10 & 29599 & Unseen Foggy weather testing \\
    ExDark& Real & 2563 & 5 & 6450 & Unseen Lowlight weather testing \\
    \bottomrule
  \end{tabular}
  \caption{Datasets.}
  \label{tab:dataset}
\end{table*}

\begin{table}[h]
\centering
\begin{tabular}{p{2cm}p{5.4cm}}
     \toprule
     \textbf{Dataset}& \textbf{Common Classes}\\
     \toprule
     PascalVOC& all\\
     RTTS& person, bicycle, car, bus, motorcycle\\
     ExDark& person, bicycle, boat, bottle, bus, car, cat, chair, dog, motorcycle \\
    \bottomrule
\end{tabular} 
\caption {Common classes with the PascalVOC dataset.}
\label{tab:classes}
\end{table}

Many of the SOTA methods for object detection in adverse weather either use the model trained on the training set of ExDark or sample the PascalVOC clear/simulated weather data to include samples with the common classes ( Table \ref{tab:classes}) present in ExDark \cite{Exdark} for low-light and RTTS \cite{RTTS} for foggy weather. We instead train or fine-tune the the WARLearn model with all the 20 classes present in the PascalVOC dataset. For a fair performance comparison, we retrain other SOTA methods on the same dataset as WARLearn. The dataset details are provided in Table \ref{tab:dataset}.

To prove the versatility of WARLearn framework, we also conduct similar experiments with the ultralytics YOLOv8 \cite{Yolov8} model on the same datasets. The clean weather training was done for 100 epochs. The starting learning rate was 0.01 for the clear weather training and was subsequently reduced by a factor to 100 to 0.0001 for the adverse weather fine tuning. The fine-tuning for YOLOv8 was also done for 10 epochs. We present these results as part of the ablation studies. 

\section{Results}

We test the WARLearn framework on adverse weather scenarios of foggy and low light conditions.

\subsection{Results with Foggy Weather Datasets}

\begin{table}[h]
\centering
\begin{tabular}{p{1.8cm}|p{1.8cm}|p{0.7cm}p{1.2cm}p{0.8cm}}
     \hline
     \textbf{Framework} & \textbf{Train Data}& \textbf{Clean} & \textbf{SimFoggy} & \textbf{RTTS} \\
     \hline
     YOLOv3 & PascalVOC train & 76.30 & 53.61 & 46.20\\
     \hline
     YOLOv3 & PascalVOC foggy mixed & 66.80 & 71.20 & 49.00\\
     \hline
     C2PNet + YOLOv3 & PascalVOC train & ---- & 66.50 & 47.70\\
     \hline
     IA-YOLO & PascalVOC foggy mixed & 69.38 & 67.58 & 47.68\\
     \hline
     DENet & PascalVOC foggy mixed & 68.40 & 65.60 & 50.50\\
     \hline
     C2PNet + WARLearn & PascalVOC foggy mixed & ---- & 69.30 & \textbf{52.00}\\
     \hline
     WARLearn & PascalVOC foggy mixed & 69.11 & \textbf{75.10} & \textbf{52.60}\\
     \hline
\end{tabular} 
\caption {Comparison of mAP@50 (\%) for various frameworks on Clean (PascalVOC test), SimFoggy (PascalVOC test foggy), and real-world foggy RTTS datasets. Last two rows are experiments with our new WARLearn framework.}
\label{tab:foggy}
\end{table}

\Cref{tab:foggy} presents the results on foggy test data, showcasing the superior performance of the WARLearn framework compared to other detection frameworks on both synthetic and real-world foggy datasets. 

The WARLearn framework surpasses end to end detectors like IA-YOLO \cite{YoloAdavtive} and DENet \cite{denet} by a huge margin. The mean Average Precision (mAP) value of 75.10\% on synthetic foggy data surpasses the closest competitor by 3.90\%. Similarly, the mAP value of 52.60\% on the RTTS \cite{RTTS} dataset exceeds the nearest competitor by 2.10\%. The drop in clear weather performance of WARLearn is minimal (0.27\% less) when compared to other frameworks trained on mixed foggy data. Even with SOTA pre-processing method like C2PNet \cite{c2pnet}, WARLearn performs better than the YOLOv3 model on both simulated foggy and real world RTTS datasets. This is due to the fact that the pre-processing does not remove all the foggy artifacts and the weights of WARLearn are learnt with mixed training data containing both clear and foggy weather images. The redundancy reduction term in the WARlearn also helps to learn more robust features.

\begin{figure}[h]
  \centering
   \includegraphics[width=1\linewidth]{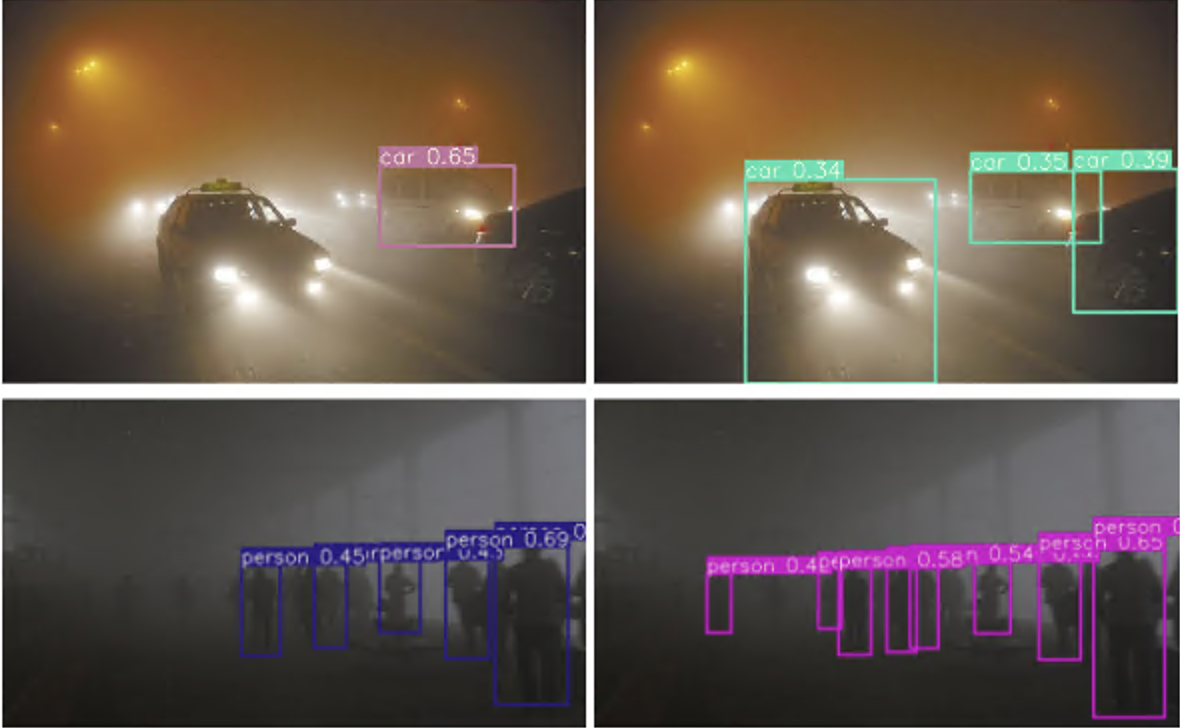}
   \caption{Comparison of detection results between YOLOv3 baseline (left column) and WARLearn (right column) on the real-world foggy RTTS dataset. WARLearn recognizes more objects in foggy conditions with higher confidence scores.}
   \label{fig:foggy}
\end{figure}

\Cref{fig:foggy} shows the performance comparison of YOLOv3 and WARLearn on a few sample images from the RTTS dataset.

\subsection{Results with Low Light Weather Datasets}

\begin{figure}[h]
  \centering
   \includegraphics[width=1\linewidth]{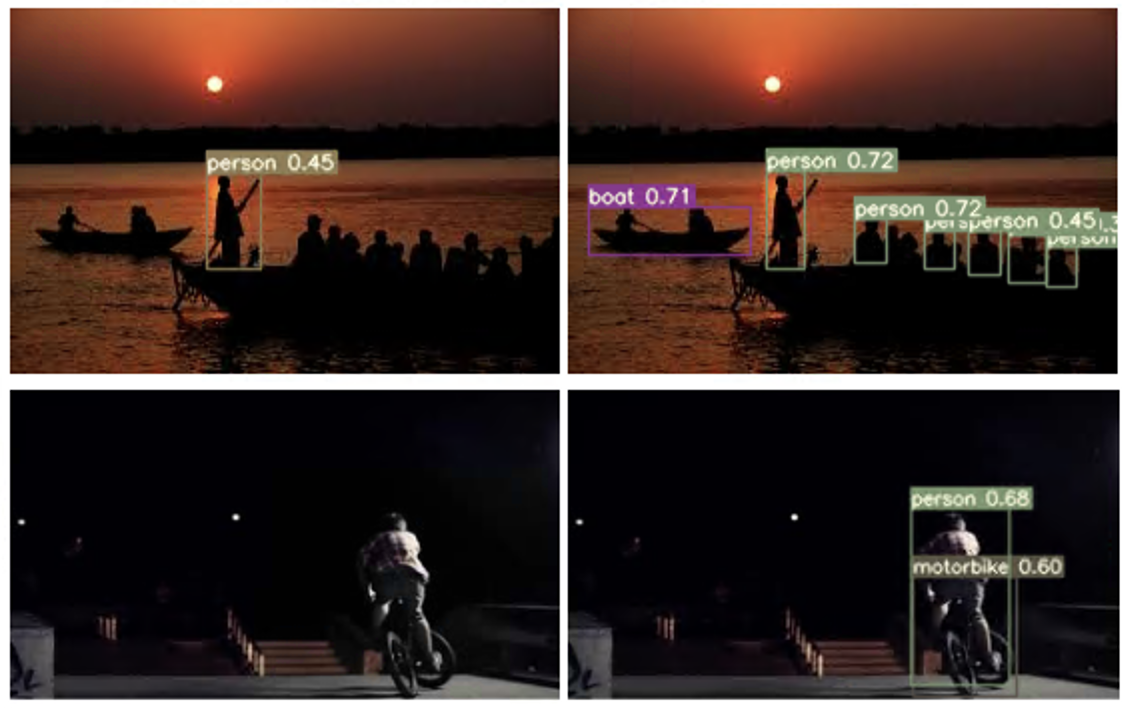}
   \caption{Comparison of detection results between YOLOv3 baseline (left column) and WARLearn (right column) on the real-world low-light ExDark dataset. WARLearn excels in recognizing more objects in low-light conditions with higher confidence scores.}
   \label{fig:lowlight}
\end{figure}

The results on low-light weather datasets mirror those obtained on foggy weather datasets. \Cref{tab:lowlight} details these results for low-light weather datasets. We obtain mAP@50 of 70.90\%  with the WARLearn framework on the synthetic low-light PascalVOC dataset and 55.70\% percent on the real-world ExDark \cite{Exdark} dataset. WARLearn surpasses the nearest mAP@50 by 2.70\% and 2.00\% on the synthetic and real world low light datasets, respectively. Here, the clear weather performance is also substantially better(3.16\% more) than other generalized frameworks trained on mixed low-light dataset. \Cref{fig:lowlight} shows the performance comparison of YOLOv3 and WARLearn on some sample images from the WARLearn dataset.

\begin{table}[h]
\centering
\begin{tabular}{p{1.8cm}|p{2.2cm}|p{0.5cm}p{0.8cm}p{0.9cm}}
     \hline
     \textbf{Framework} & \textbf{Train Data}& \textbf{Clean} & \textbf{SimLL} & \textbf{ExDark} \\
     \hline
     YOLOv3 & PascalVOC train & 76.30 & 59.90 & 49.30\\
     \hline
     YOLOv3 & PascalVOC lowlight mixed & 70.20 & 67.60 & 49.90\\
     \hline
     PairLIE + YOLOv3 & PascalVOC train & ----&60.10 & 42.70\\
     \hline
     ZeroDCE + YOLOv3 & PascalVOC train & ----&63.10 & 48.10\\
     \hline
     IA-YOLO & PascalVOC lowlight mixed & 72.34 & 63.38 & 48.16\\
     \hline
     DENet & PascalVOC lowlight mixed & 72.20 & 62.40 & 47.50\\
     \hline
     PairLIE + WARLearn & PascalVOC lowlight mixed & ---- & 63.90 & 47.00\\
     \hline
     ZeroDCE + WARLearn & PascalVOC lowlight mixed & ---- & 67.00 & 51.30\\
     \hline
     FeatEnhancer (YOLOv3) & PascalVOC lowlight mixed & ---- & 68.20 & 53.70\\
     \hline
     WARLearn & PascalVOC lowlight mixed& 75.50 & \textbf{70.90} & \textbf{55.70}\\
     \hline
\end{tabular} 
\caption {Comparison of mAP@50 (\%) for various frameworks on Clean (PascalVOC test), SimLL (PascalVOC test lowlight), and real-world low-light ExDark datasets. Last two rows are experiments with WARLearn framework.}
\label{tab:lowlight}
\end{table}

In lowlight conditions also, the SOTA pre-processing methods of ZeroDCE \cite{Zero-DCE} and PairLIE \cite{pairlie} perform better with WARLearn than the YOLOv3.  We can also observe that IA-YOLO, DENet, PairLIE and ZeroDCE+YOLOv3, do not generalize well on unseen real world foggy RTTS dataset and have a slightly lower mAP@50 compared to baseline YOLOv3. However they perform well on the synthetic PascalVOC lowlight test data that is similar to the data used for training. FeatEnHancer \cite{featenhancer} is the closest and generalises well, still falling short of WARLearn performance. This result also indicates that along with the invariance term that aligned the adverse weather features to clear features in the WARLearn, redundancy reduction loss term is also useful as for better generalization on real world data.  

In general, we note a substantial improvement in performance when utilizing the WARLearn framework compared to other state-of-the-art frameworks on adverse weather datasets. Since the WARLearn model is fine-tuned on synthetic adverse weather data, we observe a steep increase in mAP values for the synthetic adverse weather test data. This is without a significant drop in performance on clear weather dataset. Furthermore, we observe a considerable increase in mAP when applied to real-world adverse weather datasets.

\section{Ablation Studies}

\subsection{Experiments with YOLOv8}
To prove the effectiveness of the WARLearn framework across different architectures, we did the experiments with the latest YOLOv8 \cite{Yolov8} framework from ultralytics.

\begin{table}[h]
\centering
\begin{tabular}{p{1.7cm}|p{1.8cm}|p{0.6cm}p{1.2cm}p{0.9cm}}
     \hline
     \textbf{Framework} & \textbf{Train Data}& \textbf{Clean} & \textbf{SimFoggy} & \textbf{RTTS} \\
     \hline
     YOLOv8 & PascalVOC foggy mixed & 78.90 & 76.00 & 58.20\\
     \hline
     YOLOv8-WARLearn & PascalVOC foggy mixed& 79.70 & \textbf{76.60} & \textbf{59.80}\\
     \hline
\end{tabular} 
\caption {Comparison of mAP@50 (\%) for YOLOv8 and YOLOv8-WARLearn on Clean (PascalVOC test), SimFoggy (PascalVOC test foggy), and real-world foggy RTTS datasets.}
\label{tab:foggy8}
\end{table}

\begin{table}[h]
\centering
\begin{tabular}{p{1.7cm}|p{2.2cm}|p{0.6cm}p{0.8cm}p{0.9cm}}
     \hline
     \textbf{Framework} & \textbf{Train Data}& \textbf{Clean} & \textbf{SimLL} & \textbf{ExDark} \\
     \hline
     YOLOv8 & PascalVOC lowlight mixed & 81.30 & 73.60 & 57.70\\
     \hline
     YOLOv8-WARLearn & PascalVOC lowlight mixed& 79.80 & 71.00 & \textbf{60.00}\\
     \hline
\end{tabular} 
\caption {Comparison of mAP@50 (\%) for YOLOv8 and YOLOv8-WARLearn on Clean (PascalVOC test), SimLL (PascalVOC test lowlight), and real-world lowlight ExDark datasets.}
\label{tab:lowlight8}
\end{table}

Under both foggy conditions \Cref{tab:foggy8} and lowlight conditions \Cref{tab:lowlight8}, we observe significant improvements when employing the WARLearn framework with the YOLOv8 model on real world adverse weather datasets. These results demonstrate the versatility and effectiveness of the WARLearn framework, highlighting it's adaptability to various model architectures. This adaptability is not constrained by current models and can be seamlessly integrated with the latest advancements in model development.

\subsection{Parameters and Inference Time}
We conducted a comparative analysis of the inference time for various SOTA approaches alongside our proposed WARLearn method. For this comparison, we ran the inference time measurements on a 64-bit Windows machine equipped with an Intel Core i9-10900 CPU, 64 GB of RAM, and an NVIDIA GEFORCE RTX 3090 GPU with 32 GB of memory. The input image size is 416x416 for all the frameworks. For better understanding, we also compared the number of model parameters across these frameworks. The results of this analysis underscore are provided in \Cref{tab:inference}.

\begin{table}[h]
\centering
\begin{tabular}{p{2cm}|p{2.4cm}|p{2.6cm}}
     \hline
     \textbf{Framework} & \textbf{Additional Parameters}& \textbf{Additional Inference Time (in ms)} \\
     \hline
     YOLOv3 & -- & --\\
     \hline
     C2PNet & 7.16M & 77.00\\
     \hline
     PairLIE & 340K & 49.03\\
     \hline
     ZeroDCE & 79K & 6.31\\
     \hline
     IA-YOLO & 165K & 7.20\\
     \hline
     FeatEnHancer & 138K & 6.84\\
     \hline
     DENet & 45K & 5.04\\
     \hline
     WARLearn & \textbf{0} & \textbf{0.00}\\
     \hline
\end{tabular} 
\caption {Comparison of Parameters and Inference Time for different adverse weather detection frameworks. For this study, we consider YOLOv3 as a baseline and include only the additional parameteres and inference time when compared to YOLOv3.}
\label{tab:inference}
\end{table}

We can see that all the adverse weather detection frameworks including IA-YOLO \cite{YoloAdavtive} and DENet \cite{denet} require additional parameters and inference time compared to YOLOv3. C2PNet \cite{c2pnet}, PairLIE \cite{pairlie}, FeatEnHancer \cite{featenhancer}
and ZeroDCE \cite {Zero-DCE} rely on pre-processing using additional parameters. WARLearn does not introduce any additional parameters and uses the same architecture as YOLOv3, ensuring that inference time remains same as that of YOLOv3. This characteristic highlights the efficiency of WARLearn in maintaining computational performance. This shows the WARLearn's ability to deliver competitive inference times with a significant boost in performance, making it a viable option for real-time object detection tasks, especially under adverse conditions.

\subsection{Performance with Different Degradation Levels}
We also conducted experiments to observe the variation in performance across different levels of noise under foggy weather conditions. For this experiment, we divided the SimFoggy dataset into ten parts, each representing different levels of degradation with $i$ varying from 0 to 9 in \Cref{eq:beta}. This results in $\beta$ ranging from 0.05 to 0.14. Each of these ten parts contains 4952 images, with a specific $\beta$ value applied to the clear PascalVOC test dataset.

\begin{table}[h]
\centering
\begin{tabular}{p{1.8cm}|p{1.3cm}|p{0.8cm}|p{1.3cm}|p{0.8cm}}
     \hline
     \textbf{Fog Level $\beta$} & \textbf{Precision}& \textbf{Recall}& \textbf{mAP@50}& \textbf{F1}\\
     \hline
     0.05 & 0.658 & 0.779& 0.751& 0.711\\
     \hline
     0.06 & 0.641 & 0.786& 0.756& 0.704\\
     \hline
     0.07 & 0.629 & 0.799& 0.760& 0.702\\
     \hline
     0.08 & 0.621 & 0.802& 0.759& 0.697\\
     \hline
     0.09 & 0.613 & 0.807& 0.761& 0.694\\
     \hline
     0.10 & 0.602 & 0.807& 0.760& 0.687\\
     \hline
     0.11 & 0.596 & 0.806& 0.761& 0.682\\
     \hline
     0.12 & 0.584 & 0.805& 0.757& 0.673\\
     \hline
     0.13 & 0.577 & 0.800& 0.750& 0.667\\
     \hline
     0.14 & 0.566 & 0.796& 0.742& 0.658\\
     \hline
\end{tabular} 
\caption {Performace comparison of WARLearn at different levels of fog.}
\label{tab:ablation}
\end{table}

\begin{figure*}[t]
  \centering
   \includegraphics[width=1\linewidth]{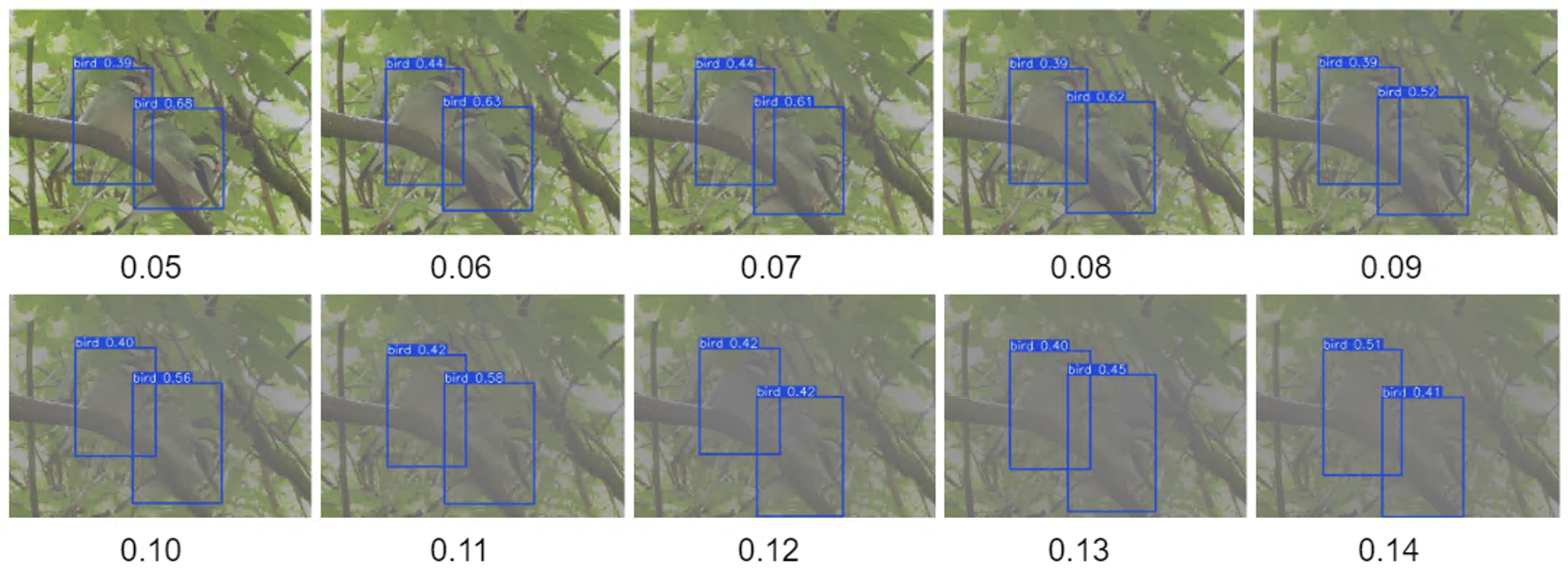}
   \caption{Detection results with WARLearn framework on an image with different levels of foggy degradation indicated by $\beta$ (\Cref{eq:beta}) value below the image. These images are taken from simulated PascalVOC test foggy dataset.}
   \label{fig:fog_levels}
\end{figure*}

As shown in \Cref{tab:ablation}, the precision and F1 score exhibit a consistent decline with increasing levels of degradation, which is expected given their sensitivity to the detection threshold. However, mAP@50 remains consistent across various levels of fog. This is further illustrated in \Cref{fig:fog_levels}, where the detection performance is similar regardless of the degradation level. There is a slight change in the bounding box at higher levels of noise, but the network consistently identifies two birds in the images, even when they are barely visible to the human eye. We observed comparable performance of WARLearn with other images as well. The few instances where WARLearn failed to detect objects occurred when the objects were far from the camera and became indistinguishable due to degradation. This analysis demonstrates that the WARLearn framework can learn robust features and maintain consistent mAP@50 performance with varying degrees of noise.

\subsection{Significance of Redundancy Reduction Term}
We conducted experiments on foggy dataset using WARLearn with $\lambda$ (Equation \ref{eq:eq2}) being 0 such that we only use the invariance term for adaptation. We can clearly see in Table \ref{tab:lambda}, that the redundancy reduction term substantially improves the WARLearn performance in adverse weather with a minor drop in performance on clear weather dataset.

\begin{table}[h]
\centering
\begin{tabular}{p{0.8cm}|p{2.5cm}|p{0.6cm}p{1.2cm}p{0.9cm}}
     \hline
     \textbf{$\lambda$} & \textbf{Condition}& \textbf{Clean} & \textbf{SimFoggy} & \textbf{RTTS} \\
     \hline
     0.000 & No redundancy reduction& 69.60 &  67.40 & 47.90\\
     \hline
     0.001 & With redundancy reduction& 69.11 &  75.10 & 52.60\\
     \hline
\end{tabular} 
\caption {Comparison of mAP@50 (\%) with and without redundancy reduction in WARLearn on Clean (PascalVOC test), SimFoggy (PascalVOC test foggy), and real-world foggy RTTS datasets. We use PascalVOC foggy mixed dataset for training these models.}
\label{tab:lambda}
\end{table}

\section{Conclusion}
WARLearn introduces a novel and efficient framework to facilitate object detection in adverse weather conditions. The experimental results validate our approach, showcasing superior performance over the existing state-of-the-art methods on both simulated and real world adverse weather datasets. The framework is also efficient and has same inference time and parameters as baseline YOLOv3. 

In our experiments, we demonstrated the effectiveness of the WARLearn framework in handling adverse conditions such as lowlight and foggy weather. However, the versatility of the WARLearn framework extends beyond these specific scenarios. As long as we can simulate an adverse condition on clean data and have a detection model with a backbone, such as YOLO, to extract the features, WARLearn can be employed to mitigate the effects of these adverse conditions and accurately detect objects. It is also well-suited for scenarios where there is a distribution shift in data over time. Future work can explore the integration of WARLearn with other state-of-the-art detection models and it's application in a broader range of challenging real-world scenarios. By doing so, we can further validate it's effectiveness, leading to development of more resilient and adaptive computer vision systems.

\section{Acknowledgment}
This research was partially supported by the Israeli Innovation Authority through the Trust.AI consortium.

{\small
\bibliographystyle{ieee_fullname}
\bibliography{egbib}
}

\end{document}